# Revealed Preference at Scale: Learning Personalized Preferences from Assortment Choices

NATHAN KALLUS, Cornell University and Cornell Tech
MADELEINE UDELL, Cornell University

We consider the problem of learning the preferences of a heterogeneous population by observing choices from an assortment of products, ads, or other offerings. Our observation model takes a form common in assortment planning applications: each arriving customer is offered an assortment consisting of a subset of all possible offerings; we observe only the assortment and the customer's single choice.

In this paper we propose a mixture choice model with a natural underlying low-dimensional structure, and show how to estimate its parameters. In our model, the preferences of each customer or segment follow a separate parametric choice model, but the underlying structure of these parameters over all the models has low dimension. We show that a nuclear-norm regularized maximum likelihood estimator can learn the preferences of all customers using a number of observations much smaller than the number of item-customer combinations. This result shows the potential for structural assumptions to speed up learning and improve revenues in assortment planning and customization. We provide a specialized factored gradient descent algorithm and study the success of the approach empirically.

## 1. INTRODUCTION

In many commerce and e-commerce settings, a firm chooses a set of items (products, ads, or other offerings) to present to a customer, who then chooses from among these items. Each choice results in some revenue for the firm that depends on the item selected. The problem of choosing the revenue-maximizing assortments of items to offer to the customer is referred to as *assortment planning* or *assortment optimization*. To solve this problem, a firm must estimate customer preferences, and then choose an optimal assortment of goods to present based on those estimates.

Usually the number of interactions between the firm and customer is limited so efficient estimation of customer preferences is critical. But estimating customer preferences is no easy task: there are combinatorially many assortments of goods, and so potentially combinatorially many quantities to estimate. To enable tractable estimation, customer preferences are often modeled parametrically, using the multinomial logit (MNL) model or its variants.

However, one parameter vector may not fit everyone in a *heterogeneous* population. For a better fit, it can be important to *segment* the population into groups (geographic, demographic, or temporal) and to fit parameters for each segment. In the limit, each customer may represent a separate segment. In the e-commerce and online advertising settings, populations can be segmented into individual customers, since firms have data on the indidual customers' choices from individually customized assortments. In the offline brick-and-mortar retail setting, populations may be segmented by store branch, since firms have data on aggregate choices in each store branch and tastes often vary geographically and by venue (mall, street, etc.). Customers may also be segmented into smaller groups using loyalty program data.

The number of observations needed to estimate the parameters of the model for each segment is at least linear in the number of items. But in modern advertisement markets or online retail settings, tens of millions of items may be on offer: far more than the number of ads any one customer can be expected to view or click, or products any one customer can be expected to consider or buy. Moreover, the sort of data available in practice is limited to observation of a single choice of a single item out of a whole assortment, rather than a full ranking or a pairwise comparison.

In this paper we propose a new model that enables tractable estimation as the number of segments and items grows. We suppose the preferences of each type (individual customer or customer segment) follow a parametric (MNL) choice model, while the underlying structure of these parameters over types has a low dimension. We show that a nuclear-norm regularized maximum likelihood estimator based on such data can learn the preferences of all customers using a number of observations that grows sublinearly in the number of type-item combinations. This result shows the potential for structural assumptions to speed up learning and improve revenue in assortment planning and customization.

### 1.1. Related Work

The model presented in this paper builds on two distinct lines of research: on choice modelling and assortment optimization, and on low-rank matrix completion.

*Choice Modelling and Assortment Optimization.* Assortment optimization is a central problem in revenue management: which items should be presented to a given customer in order to maximize the expected revenue from the customer's choice?

The first step in answering this question is to understand how customers choose from among a set of goods. Discrete choice models posit answers to this question in the form of a probability distribution over choices. Luce [1959] proposed an early discrete choice model based on an axiomatic theory, resulting in the basic attraction model. Later, the work of McFadden [1973] on random utility theory led to the MNL model, which posits that customer choices follow a log-linear model in a vector of customer preference parameters. Fitting a single MNL model is as simple as counting the number of times an item is chosen relative to the other offerings. (These counts give the maximum likelihood estimate for the model.) Under the MNL model, it is easy to optimize assortments: presenting items in revenue sorted order is optimal [Talluri and Van Ryzin 2006].

Conceiving of assortments as arms and consumer choice as bandit feedback, Rusmevichientong et al. [2010] and Sauré and Zeevi [2013] consider dynamic assortment optimization under a single MNL model. The former also present a polynomial-time algorithm for optimizing assortment under an MNL model and with cardinality constraints.

The mixture of MNLs (MMNL) model models consumer choice as a mixture of MNL models with different parameters. Unfortunately, it is NP-hard to optimize a single assortment to be offered to an MMNL population, even with only two mixture components [Rusmevichientong et al. 2014]. Other derivatives of the MNL model include the nested logit model [Williams 1977] and its extensions [McFadden 1980]. Assortment optimization over these is also computationally hard [Davis et al. 2014].

*Matrix Completion.* Recent years have seen a surge of interest in matrix completion: the problem of (approximately) recovering an (approximately) low rank matrix from a few (noisy) samples from its values. The surprising result is that simple algorithms, such as nuclear norm regularized maximum likelihood estimation, can often recover a low rank matrix given only a small number of observations in each row or column.

Following groundbreaking work on exact completion of exactly low rank matrices whose entries are observed without noise [Candès and Recht 2009; Candès and Tao 2010; Keshavan et al. 2010; Recht et al. 2010], approximate recovery results have been obtained for a variety of different noisy observation models. These include observations with additive gaussian [Candès and Plan 2009] and subgaussian [Keshavan et al. 2009a] noise, 0-1 (Bernoulli) observations [Davenport et al. 2012], observations from any exponential family distribution [Gunasekar et al. 2014], and observations generated according to the Bradley-Terry-Luce model for pairwise comparisons [Lu and Negahban 2014].

Our results follow in the vein of the statistical matrix completion bounds. The proof of our main result uses the machinery of restricted strong convexity developed by Negahban and Wainwright [2012], and echoes many of the ideas in the technical report by Lu and Negahban [2014], which proved a similar sample complexity result for matrix completion from observations of pairwise ranks. Our method extends these ideas to address observations consisting only of a single choice from an arbitrarily-sized subset of all items. Most recently (and independently of our work), Oh et al. [2015] extended the results of Lu and Negahban [2014] to observations of the ranking of all items in a subset, rather than the (pairwise) ranking of items in a subset of size two. A primary distinguishing feature of our work is the observation model we consider — observing choices, rather than rankings — which applies naturally to the type of data available in realistic applications of assortment planning. This type of data is much more common in practice because it corresponds to passive observations of consumer behavior that is truthful (assuming choice is utility maximizing) and it is available in large amounts because it can be derived from transactional data. We also use weaker assumptions on the distribution of assortment sets offered: in particular, Oh et al. [2015] require that the sets contain duplicate items (with nonzero probability), where duplicated items are chosen with higher frequency, violating independence of irrelevant alternatives.

**1.2. Contributions**

This paper makes four main contributions.

— We propose the *low rank MMNL* model for customer preferences: the preferences of each type follow a parametric (MNL) choice model, but the underlying latent structure of parameters over types has low dimension.
— We consider the problem of learning such a choice model from observations of choices from assortments and propose a nuclear-norm regularized maximum likelihood estimator.
— Theorem 3.1 provides the first bound on sample complexity of learning this model from choice data.
— Algorithm 1 provides a fast method for computing our estimator with a small memory footprint.

## 2. PROBLEM STATEMENT AND ALGORITHM

We suppose that at each time $t = 1, \ldots, T$ a customer arrives with type $i_t$ chosen uniformly at random from the set $\{1, \ldots, m\}$. The customer is presented with a choice of items $S_t \subseteq \{1, \ldots, j\}$ of size $|S_t| = K_t$, where $S_t|K_t$ is sampled uniformly from the set of subsets of $\{1, \ldots, n\}$ of size $K_t$. We make no assumption on the distribution of $K_t$ other than that $K_t \leq K$ is bounded almost surely.

The customer then chooses an item according to a multinomial logit model: item $j_t \in S_t$ is chosen with probability

$$\mathbb{P}\left(j_t = j \mid i_t = i, S_t = S\right) = \frac{e^{-\Theta^\star_{ij}}}{\sum_{j' \in S} e^{-\Theta^\star_{ij'}}}. \tag{1}$$

The standing assumption is that $\Theta^\star$ is of low underlying dimension, either having low rank $r \ll m, n$ or approximately low rank (see below for details).

After $T$ observations $(i_t, j_t, S_t)$ from this model, we wish to estimate the parameter matrix $\Theta^\star$. To eliminate redundant degrees of freedom, we assume without loss of generality that $\sum_{j=1}^n \Theta^\star_{ij} = 0$ for every $i = 1, \ldots, m$, i.e., $\Theta^\star e = 0$. We also assume that $||\Theta^*||_\infty \leq \alpha/\sqrt{mn}$ for purely technical reasons; see below. This assumption is standard in matrix completion recovery results.

*Discussion.* Model (1) describes an MMNL over mixture components indexed by $i$. Using the approximation results of McFadden and Train [2000], for any choice distribution over a population, there is a variable $I$ such that, the choice distribution is approximately MNL conditioned on $I$. Hence the model above can represent *any* choice distribution so long as the population is segmented finely enough.

The MMNL model is equivalent to a utility maximizing model under a particular distribution of customer utilities. Define $u_{ij} = -\Theta^\star_{ij}$ to be the nominal utility of type $i$ for item $j$. Suppose that a random customer of type $i$ has utility $u_{ij} + \zeta_{ij}$, where $\zeta_{ij} \sim \text{Gumbell}(0, 1)$ is a random idiosyncratic deviation from the nominal utility distributed according to the extreme value distribution. When presented with an assortment, each customer simply chooses the product $j_t$ maximizing her own utility function:

$$j_t = \max_{j \in S_t} \left(u_{ij} + \zeta_{ij}\right). \tag{2}$$

Our mathematical model (1) can be applied in two conceptually distinct problem settings. In the first setting, each type $i$ represents a group of customers, each one of which has an idiosyncratic utility distributed as in (2). We say that the population is heterogeneous because each type $i$ is associated with a different nominal utility $u_{ij}$ for each product $j$. In the second setting, each type $i$ represents a single customer. The random idiosyncrasies associated with each choice made by the same customer reflect human inconsistencies in decision making, or slight variations in preferences over time [DeShazo and Fermo 2002; Kahneman and Tversky 1979].

The observation model we propose has several advantages. Our observations consist of customers' choices from assortments. These choices are typically the only observation possible in applications of assortment planning. Moreover, these observations tend to be truthful: customers generally make choices to maximize their utility. Observed choices should be contrasted with data obtained by interrogating customers about their ranking of products (for example, in surveys or focus groups); these stated preferences are known to be unreliable as indicators of true, revealed preferences [Samuelson 1948].

One limitation of our results is that we require a somewhat random design (random $S_t$) to guarantee that preferences are learned accurately. Hence our results can either be interpreted as a prescription on how to design practical, consistent experiments in consumer choice where only choices are observed, or they can be interpreted as theoretical justification for our algorithm, even if applied to data of general design. Indeed, the empirical success of matrix completion approaches on real world data sets suggests that the algorithm may work well even when applied to non-random observations [Funk 2006]. Other authors have suggested variations on the nuclear norm regularizer we propose to compensate for non-uniform sampling distributions [Foygel et al. 2011]; generalizing these variations to observations of assortments is an interesting problem beyond the scope of this paper.

We are not the first to model customer preferences as a mixture of MNLs. For example, Bernstein et al. [2011] consider the problem of multi-stage assortment optimization over time with limited inventories and for consumers from two or more segments, each distributed as MNL. They do not consider a learning problem but consider a stimulation study of their optimization techniques based on MNL distributions estimated from real choice data. This estimation method scales poorly with the dimension of the problem and so is limited to very few segments (three in their case study). Our model, on the other hand, scales easily to very large problems: fixing the rank of the parameter matrix, the complexity of the estimation problem increases linearly with the number of types and products, rather than the number of their combinations.

*Algorithm.* Define the negative log likelihood of the observations given parameter $\Theta$ as

$$L_T(\Theta) = \frac{1}{T} \sum_{t=1}^{T} \log \left( \sum_{j \in S_t} e^{\Theta_{i_t j_t} - \Theta_{i_t j}} \right).$$

Define the estimator $\widehat{\Theta}$ to be any solution of the nuclear norm regularized maximum likelihood problem

$$\begin{aligned}
\text{minimize} \quad & L_T(\Theta) + \lambda \|\Theta\|_*, \\
\text{subject to} \quad & \Theta e = 0, \\
& \|\Theta\|_\infty \leq \alpha/\sqrt{mn},
\end{aligned} \quad (3)$$

where $\lambda > 0$ is a parameter, $e$ is the vector of all 1's, and we define the *nuclear norm* $\|\Theta\|_*$ to be the sum of the singular values of $\Theta$. The constraint $\|\Theta\|_\infty \leq \alpha/\sqrt{mn}$ appears as an artifact of the proof; this last constraint can be omitted without sacrificing good practical performance. See Sec. 6.

Problem (3) is convex and hence can be solved by a variety of standard methods [Boyd and Vandenberghe 2004]. In Sec. 5 we provide a specialized first-order algorithm that works on the *non-convex*, factored form of the problem.

## 3. MAIN RESULT

We bound the error of our estimator (3) in terms of the following quantities, which capture the complexity of learning the preferences of all customer types over all items.

— *Number of observations.* The bound decreases as the number of observations increases.
— *Number of parameters.* The bound grows as $d = \frac{m+n}{2}$ grows, where $\Theta^*$ consists of $m \times n$ parameters.
— *Underlying rank dimension.* For any $r < \min(m, n)$, our bound decomposes into two error terms. The first error term is the error in estimating the top $r$ "principal components" of the parameter matrix. This error grows with $\sqrt{r}$ and captures the benefit of learning only the most salient features instead of all parameters at once. The second error is the error in approximating the parameter matrix by only its top $r$ "principal components." In particular, if it is assumed that $\Theta^\star$ is exactly rank $r$ then this latter error is always zero. More generally, however, we may conceive of parameter matrices that are approximately low rank, i.e., that have quickly decaying singular values past the top $r$, which would lead to a nonzero but small approximation error.
— *Size of parameters*. Our bound grows with the (scaled) maximum magnitude of any entry $\alpha$.
— *Size of choice sets*. Our bound grows with the maximum size of the choice sets $K$.

THEOREM 3.1. *Suppose $T \leq d^2 \log d$ and $\lambda = 32\sqrt{\frac{Kd\log d}{mnT}}$. Then under the observation model described and for any integer $r \leq \min(m,n)$, with probability at least $1 - 4/d^3$, any solution $\widehat{\Theta}$ to Problem (3) satisfies*

$$\|\widehat{\Theta} - \Theta^\star\|_F \leq 2048\alpha e^{\frac{6\alpha}{\sqrt{mn}}} \max\left\{\sqrt{\frac{K^3 d \log(d)}{T}}\sqrt{r},\right.$$

$$\left.\left(\frac{K^3 d \log(d)}{T}\right)^{1/4} \left(\sum_{j=r+1}^{\min\{m,n\}} \sigma_j(\Theta^\star)\right)^{1/4}\right\}.$$

## 4. PROOF OF MAIN RESULT

Define $\Delta = \widehat{\Theta} - \Theta^\star$ as the (matrix) error to bound, $K_t = |S_t|$, $S'_t = S_t \setminus \{j_t\}$, $\gamma = \alpha/\sqrt{mn}$, $X_{tj} = e_{i_t} e_{j_t}^T - e_{i_t} e_j^T$, and $Y_t(\Theta) = \text{Var}(\{\Theta_{i_t j} : j \in S_t\}) = \frac{1}{K_t} \sum_{j \in S_t} \left(\Theta_{i_t j} - \frac{1}{K_t} \sum_{j' \in S_t} \Theta_{i_t j'}\right)^2$. We will use the following three lemmas, whose proofs we defer to Sec. 8.

LEMMA 4.1. *Let*

$$\mathcal{A}_{\Gamma,\nu} = \left\{\Theta : \|\Theta\|_\infty \leq \gamma, \|\Theta\|_F \leq \Gamma, \|\Theta\|_* \leq \frac{\nu}{240\sqrt{Kmn\gamma}}\sqrt{\frac{T}{d\log d}}\Gamma^2, \Theta e = 0\right\}$$

*and*

$$\mathcal{M}_{\Gamma,\nu} = \sup_{\Theta \in \mathcal{A}_{\Gamma,\nu}} \left(\frac{1}{m}\frac{1}{n}\|\Theta\|_F^2 - \frac{1}{T}\sum_{t=1}^T Y_t(\Theta)\right).$$

*Then*

$$\mathbb{P}\left(\mathcal{M}_{\Gamma,\nu} \geq \nu \frac{\Gamma^2}{mn}\right) \leq \exp\left(-\frac{8}{9}\frac{\nu^2}{m^2 n^2}\frac{\Gamma^4}{\gamma^4}T\right).$$

LEMMA 4.2. *Let*

$$\mathcal{A}^\star = \left\{\Theta : \|\Theta\|_\infty \leq \gamma, \|\Theta\|_* \leq \frac{1}{128\sqrt{Kmn\gamma}}\sqrt{\frac{T}{d\log d}}\|\Theta\|_F^2, \Theta e = 0\right\}.$$

*Then*

$$\mathbb{P}\left(\frac{1}{T}\sum_{t=1}^T Y_t(\Theta) \geq \frac{1}{2mn}\|\Theta\|_F^2 \ \forall \Theta \in \mathcal{A}^\star\right) \geq 1 - 2d^{-2^{27}}.$$

LEMMA 4.3. *With probability at least $1 - 2d^{-3}$,*

$$\|\nabla L_T(\Theta^\star)\|_2 \leq 16\sqrt{\frac{Kd\log d}{mnT}}.$$

PROOF (THEOREM 3.1). Let us first assume $\Delta \in \mathcal{A}^\star$, and restrict to the high probability event that the events in both Lemma 4.2 and Lemma 4.3 occur.

Define $D = L_T(\widehat{\Theta}) - L_T(\Theta^\star) - \nabla L_T(\Theta^\star) \cdot (\widehat{\Theta} - \Theta^\star)$. By Taylor's theorem, $\exists s \in [0,1]$ such that

$$D = \nabla^2 L_T(\Theta^\star + s\Delta)[\Delta, \Delta]$$

$$= \frac{1}{T} \sum_{t=1}^{T} \left( \frac{(1 + \sum_{j \in S'_t} e^{v_{tj}})(\sum_{j \in S'_t} e^{v_{tj}} (X_{t_j} \cdot \Delta)^2)}{(1 + \sum_{j \in S'_t} e^{v_{tj}})^2} - \frac{(\sum_{j \in S'_t} e^{v_{tj}} X_{t_j} \cdot \Delta)^2}{(1 + \sum_{j \in S'_t} e^{v_{tj}})^2} \right)$$

$$\geq \frac{1}{T} \sum_{t=1}^{T} \frac{1}{(1 + \sum_{j \in S'_t} e^{v_{tj}})^2} \sum_{j \in S'_t} e^{v_{tj}} (X_{tj} \cdot \Delta)^2$$

where $v_{tj} = X_{tj} \cdot (\Theta^\star + s\Delta)$ and the last inequality is Jensen's. Since $\|\Theta^\star\|_\infty, \|\widehat{\Theta}\|_\infty \leq \gamma$ we have $|v_{tj}| \leq 2\gamma$ and since the mean minimizes sum of squared distances,

$$D \geq \frac{1}{T} \frac{1}{e^{6\gamma}} \sum_{t=1}^{T} \frac{1}{K_t^2} \sum_{j \in S'_t} (X_{tj} \cdot \Delta)^2 \geq \frac{1}{T} \frac{1}{K} \frac{1}{e^{6\gamma}} \sum_{t=1}^{T} Y_t(\Delta). \tag{4}$$

Let $\Theta^\star = U \operatorname{diag}(\sigma_1, \sigma_2, \dots) V^T$ be the singular-value decomposition (SVD) of $\Theta^\star$ with singular values sorted largest to smallest. Using block notation and following Recht et al. [2010], let us rewrite $\Delta$ and define $\Delta', \Delta''$

$$U^T \Delta V = \Gamma = \begin{pmatrix} \Gamma_{11} & \Gamma_{12} \\ \Gamma_{21} & \Gamma_{22} \end{pmatrix} \text{ with } \Gamma_{11} \in \mathbb{R}^{r \times r}$$

$$\Delta'' = U \begin{pmatrix} 0 & 0 \\ 0 & \Gamma_{22} \end{pmatrix} V^T, \quad \Delta' = U \begin{pmatrix} \Gamma_{11} & \Gamma_{12} \\ \Gamma_{21} & 0 \end{pmatrix} V^T.$$

Then $\Delta = U\Gamma V^T = \Delta' + \Delta''$. Note that,

$$\operatorname{rank}(\Delta') = \operatorname{rank}(U^T \Delta' V) = \operatorname{rank}\left( \begin{pmatrix} \Gamma_{11}/2 & \Gamma_{12} \\ 0 & 0 \end{pmatrix} + \begin{pmatrix} \Gamma_{11}/2 & 0 \\ \Gamma_{21} & 0 \end{pmatrix} \right) \leq 2r.$$

Letting $\Theta_r^\star = U \operatorname{diag}(\sigma_1, \dots, \sigma_r, 0, 0, \dots) V^T$ and its complement $\overline{\Theta}_r^\star = \Theta^\star - \Theta_r^\star$, we see

$$\|\widehat{\Theta}\|_* = \|\Theta^\star + \Delta\|_* = \|\Theta_r^\star + \overline{\Theta}_r^\star + \Delta' + \Delta''\|_*$$
$$\geq \|\Theta_r^\star + \Delta''\|_* - \|\overline{\Theta}_r^\star\|_* - \|\Delta'\|_*$$
$$= \|\Theta_r^\star\|_* + \|\Delta''\|_* - \|\overline{\Theta}_r^\star\|_* - \|\Delta'\|_*$$
$$= \|\Theta^\star\|_* + \|\Delta''\|_* - 2\|\overline{\Theta}_r^\star\|_* - \|\Delta'\|_*,$$

and so $\|\widehat{\Theta}\|_* - \|\Theta^\star\|_* \leq 2\|\overline{\Theta}_r^\star\|_* + \|\Delta'\|_* - \|\Delta''\|_*$.

By the optimality of $\widehat{\Theta}$, we have

$$L_T(\widehat{\Theta}) + \lambda \|\widehat{\Theta}\|_* \leq L(\Theta^\star) + \lambda \|\Theta^\star\|_*.$$

Hence, by Hölder's inequality,

$$0 \leq D = L_T(\widehat{\Theta}) - L_T(\Theta^\star) - \nabla L_T(\Theta^\star) \cdot \Delta$$
$$\leq \|\nabla L_T(\Theta^\star)\|_2 \|\Delta\|_* + \lambda \left( \|\Theta^\star\|_* - \|\widehat{\Theta}\|_* \right). \tag{5}$$

Since $\|\nabla L_T(\Theta^\star)\|_2 \leq \lambda$, triangle inequality in (5) yields

$$D \leq \|\nabla L_T(\Theta^\star)\|_2 \|\Delta\|_* + \lambda \|\Delta\|_* \leq 2\lambda \|\Delta\|_*.$$

Together with Lemma 4.2 and eq. (4) (for the lower bound) and our choice of $\lambda$ (for the upper bound), this yields

$$\frac{1}{2e^{6\gamma}mnK}\|\Delta\|_{\mathrm{F}}^2 \leq D \leq 64\sqrt{\frac{Kd\log d}{mnT}}\|\Delta\|_*. \tag{6}$$

Hence, recalling $\gamma = \frac{\alpha}{\sqrt{mn}}$,

$$\|\Delta\|_{\mathrm{F}}^2 \leq 128\alpha e^{\frac{6\alpha}{\sqrt{mn}}} K^{3/2} \sqrt{\frac{d\log d}{T}}\|\Delta\|_*. \tag{7}$$

Returning to (5), since $\|\nabla L_T(\Theta^\star)\|_2 \leq \lambda/2$, we have

$$0 \leq \|\nabla L_T(\Theta^\star)\|_2 \|\Delta\|_* + \lambda\left(\|\Theta^\star\|_* - \|\widehat{\Theta}\|_*\right)$$

$$\leq \lambda\left(\frac{1}{2}\|\Delta\|_* + 2\|\overline{\Theta}_r^\star\|_* + \|\Delta'\|_* - \|\Delta''\|_*\right)$$

$$\leq \lambda\left(2\|\overline{\Theta}_r^\star\|_* + \frac{3}{2}\|\Delta'\|_* - \frac{1}{2}\|\Delta''\|_*\right),$$

and so $\|\Delta''\|_* \leq 3\|\Delta'\|_* + 4\|\overline{\Theta}_r^\star\|_*$. Let $\rho = \|\overline{\Theta}_r^\star\|_* = \sum_{j=r+1}^{\min\{m,n\}} \sigma_j$. Since $\|\Delta\|_{\mathrm{F}}^2 - \|\Delta'\|_{\mathrm{F}}^2 = \|\Gamma_{22}\|_{\mathrm{F}}^2 \geq 0$,

$$\|\Delta\|_* \leq \|\Delta'\|_* + \|\Delta''\|_* = 4\|\Delta'\|_* + 4\rho$$
$$\leq 8\max\{\|\Delta'\|_*, \rho\} \leq 8\max\left\{\sqrt{2r}\|\Delta'\|_{\mathrm{F}}, \rho\right\}$$
$$\leq 16\max\left\{\sqrt{r}\|\Delta\|_{\mathrm{F}}, \rho\right\}. \tag{8}$$

If $\sqrt{r}\|\Delta\|_{\mathrm{F}} \geq \rho$, substitute (8) in (7) to see

$$\|\Delta\|_{\mathrm{F}} \leq 2048\alpha e^{\frac{6\alpha}{\sqrt{mn}}} K^{3/2} \sqrt{\frac{rd\log d}{T}}.$$

Otherwise (if $\sqrt{r}\|\Delta\|_{\mathrm{F}} < \rho$), substitute (8) in (7) and take the square root to see

$$\|\Delta\|_{\mathrm{F}} \leq \sqrt{2048\alpha e^{\frac{6\alpha}{\sqrt{mn}}} K^{3/2} \sqrt{\frac{\rho d\log d}{T}}}$$
$$\leq 2048\alpha e^{\frac{6\alpha}{\sqrt{mn}}} K^{3/4} \left(\frac{\rho d\log d}{T}\right)^{1/4}.$$

Combining yields the statement.

Our last step is to investigate the case where $\Delta \notin \mathcal{A}^\star$, i.e.,

$$\|\Delta\|_* \geq \frac{1}{128\sqrt{Kmn\gamma}}\sqrt{\frac{T}{d\log d}}\|\Delta\|_{\mathrm{F}}^2.$$

In this case we can't use (7), which relies on Lemma 4.2. But rewriting and introducing redundant terms greater than 1, we see

$$\|\Delta\|_{\mathrm{F}}^2 \leq 128\alpha e^{\frac{6\alpha}{\sqrt{mn}}} K^{3/2} \sqrt{\frac{d\log d}{T}}\|\Delta\|_*,$$

Hence we recover (7) whether or not $\Delta \notin \mathcal{A}^\star$. $\square$

**ALGORITHM 1:** Factored Gradient Descent for (9)
---
**input:** dimensions $m$, $n$, $\tilde{r}$, data $\{(i_t, j_t, S_t)\}_{t=1}^T$, regularizing coefficient $\lambda$, and tolerance $\tau$
$U \leftarrow U^0$, $V \leftarrow V^0$, $f' \leftarrow \infty$.
**repeat**
  $\eta \leftarrow 1$, $f \leftarrow f'$, $\Delta U \leftarrow -\lambda U$, $\Delta V \leftarrow -\lambda V$
  **for** $t = 1, \ldots, T$ **do**
    $Q = 0$
    **for** $j \in S_t$ **do**
      $w_j \leftarrow e^{-U_{i_t}^T V_j}$
      $W \leftarrow W + w_j$
    **end for**
    $\Delta U \leftarrow \Delta U - \frac{1}{T} \left( e_{i_t} V_{j_t}^T - \frac{1}{W} \sum_{j \in S_t} w_j e_{i_t} V_j^T \right)$
    $\Delta V \leftarrow \Delta V - \frac{1}{T} \left( e_{j_t} U_{i_t}^T - \frac{1}{W} \sum_{j \in S_t} w_j e_j U_{i_t}^T \right)$
  **end for**
  **repeat**
    $U' \leftarrow U + \eta \Delta U$, $V' \leftarrow V + \eta \Delta V$
    $f' \leftarrow L_T(U'V'^T) + \frac{\lambda}{2} \|U'\|_F^2 + \frac{\lambda}{2} \|V'\|_F^2$
    $\eta \leftarrow \beta_{\text{dec}} \eta$
  **until** $f' \leq f$
  $U \leftarrow U'$, $V \leftarrow V'$
**until** $\frac{f - f'}{f'} \leq \tau$
**output:** $UV^T - UV^T ee^T / n$

## 5. A FACTORED GRADIENT DESCENT ALGORITHM

In this section we provide a factored gradient descent (FGD) algorithm for the problem

$$\begin{aligned} &\text{minimize} && L_T(\Theta) + \lambda \|\Theta\|_*, \\ &\text{subject to} && \Theta e = 0. \end{aligned} \quad (9)$$

As discussed in Sec. 2, the algorithm we employ does *not* enforce the constraint $\|\Theta\|_\infty$, as is common for matrix completion. This constraint is necessary for the technical result in our main theorem, but is unnecessary in practice, as can be seen in our numerical results in Sec. 6.

In applications, one is interested in solving the problem (9) for very large $m$, $n$, $T$. Due to the complexity of Cholesky factorization, this rules out theoretically-tractable second-order interior point methods. One standard approach is to use a first-order method, such as Cai et al. [2010]; Hazan [2008]; Orabona et al. [2012]; Parikh and Boyd [2014]; however, this approach requires (at least a partial) SVD at each step. An alternative approach, which we take here, is to optimize as variables the factors $U \in \mathbb{R}^{m \times \tilde{r}}$ and $V \in \mathbb{R}^{n \times \tilde{r}}$ of the optimization variable $\Theta = UV^T$ rather than producing these via SVD at each step; see, *e.g.*, Jain et al. [2013]; Keshavan et al. [2009b]. To guarantee equivalence of the problems, we must take $\tilde{r} = \min(m, n)$. However, if we believe the solution is low rank, we may use a smaller $\tilde{r}$, reducing computational work and storage.

Our FGD algorithm proceeds by applying gradient descent steps to the unconstrained problem

$$\begin{aligned} &\text{minimize} && L_T(UV^T) + \frac{\lambda}{2} \|U\|_F^2 + \frac{\lambda}{2} \|V\|_F^2, \\ &\text{subject to} && U \in \mathbb{R}^{m \times \tilde{r}}, V \in \mathbb{R}^{n \times \tilde{r}}. \end{aligned} \quad (10)$$

LEMMA 5.1. *Problem* (10) *is equivalent to Problem* (9) *subject to the additional constraint* $\text{rank}(\Theta) \leq \tilde{r}$.

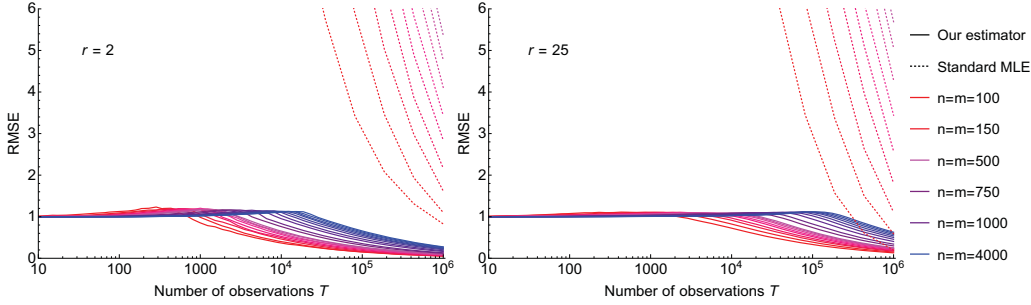

Fig. 1: RMSE of estimators for $\Theta^\star$ with $m = n = 100, 150, 200, \ldots, 500, 750, \ldots, 4000$ and $r = 2, 25$.

PROOF. Given $\Theta$ feasible in (9) with $\text{rank}(\Theta) \leq \tilde{r}$, write its SVD $\Theta = \tilde{U}\Sigma\tilde{V}^T$, where $\Sigma \in \mathbb{R}^{\tilde{r}\times\tilde{r}}$ is diagonal and $U$, $V$ unitary. Letting $U = \tilde{U}\Sigma^{1/2}$ and $V = \tilde{V}\Sigma^{1/2}$, we obtain a feasible solution to (10) with the same objective value $\Theta$ has in (9). Conversely, given $U$, $V$ feasible in (10), let $\Theta = UV^T - UV^T ee^T/n$. Note $\text{rank}(\Theta) \leq \tilde{r}$, $\Theta e = 0$, $L_T(UV^T) = L_T(\Theta)$ as $L_T(\cdot)$ is invariant to constant shifts to rows, and

$$\|\Theta\|_* \leq \|UV^T\|_* = \text{tr}(\Sigma) = \text{tr}(\tilde{U}^T U V^T \tilde{V})$$
$$\leq \|\tilde{U}^T U\|_\text{F} \|V^T \tilde{V}\|_\text{F} \leq \|U\|_\text{F} \|V\|_\text{F} \leq \frac{1}{2}\|U\|_\text{F}^2 + \frac{1}{2}\|V\|_\text{F}^2.$$

Hence, $\Theta$ has objective value no worse than $(U, V)$. □

It is easy to compute the the gradients of the objective of (10). Since $L_T(\Theta)$ is differentiable,

$$\nabla_U L_T(UV^T) = \nabla L_T(UV^T) V,$$
$$\nabla_V L_T(UV^T) = \nabla L_T(UV^T)^T U.$$

We do not need to explicitly form $\nabla L_T(UV^T)$ in order to compute these; this observation reduces the memory required to implement the algorithm (see Algorithm 1). Similarly, we need not form $UV^T$ to compute $L_T(UV^T)$. Recent work has shown that gradient descent on the factors converges linearly to the global optimum for problems that enjoy restricted strong convexity [Bhojanapalli et al. 2015]. In eq. (6) we establish restricted strong convexity for our problem with high probability.

We initialize FGD using a technique recommended by Bhojanapalli et al. [2015] which only requires access to gradients of the objective of (9). Using the SVD, write $-\nabla L_T(0) = \tilde{U} \text{diag}(\tilde{\sigma}_1, \ldots, \tilde{\sigma}_{\min(m,n)})\tilde{V}^T$ and initialize

$$U^0 = \gamma^{-1/2} \text{diag}(\sqrt{\tilde{\sigma}_1}, \ldots, \sqrt{\tilde{\sigma}_{\tilde{r}}})\tilde{U}_{:,\tilde{r}},$$
$$V^0 = \gamma^{-1/2} \text{diag}(\sqrt{\tilde{\sigma}_1}, \ldots, \sqrt{\tilde{\sigma}_{\tilde{r}}})\tilde{V}_{:,\tilde{r}},$$

where $\gamma = \|\nabla L_T(0) - (\nabla L_T(e_1 e_1^T) + \lambda e_1 e_1^T)\|_\text{F}$ and $\tilde{U}_{:,\tilde{r}}$, $\tilde{V}_{:,\tilde{r}}$ denote the first $r$ columns of $\tilde{U}$, $\tilde{V}$. We use an adaptive step size with a line search that guarantees descent. Starting with a stepsize of $\eta = 1$, the stepsize is repeatedly decreased by a factor $\beta_{\text{dec}}$ until the step produces a decrease in the objective. We terminate the algorithm when the decrease in the relative objective value is smaller than the convergence tolerance $\tau$.

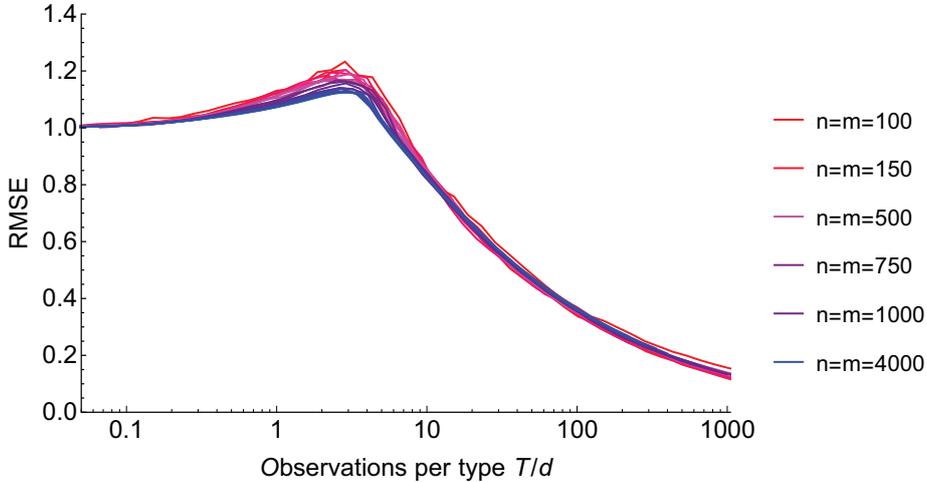

Fig. 2: RMSE of our estimators for $\Theta^\star$ by observation per row with $r = 2$.

## 6. EXPERIMENTAL RESULTS

In this section we study the problem experimentally to investigate the success of our algorithm. We compare our estimate $\widehat{\Theta}$ with the standard maximum likelihood estimate $\widehat{\Theta}^{\text{MLE}}$ that solves

$$\begin{aligned} \text{minimize} \quad & L_T(\Theta), \\ \text{subject to} \quad & \Theta e = 0. \end{aligned} \qquad (11)$$

Note that since it imposes no structure on the whole matrix $\Theta$, problem (11) decomposes into $m$ subproblems for each type (row of $\widehat{\Theta}^{\text{MLE}}$), each solving a separate MNL MLE in $n$ variables. In our experiments, we use Newton's method as implemented by *Optim.jl* to solve each of the subproblems, omitting the constraint $\Theta e = 0$ and projecting onto it at termination since $L_T(\cdot)$ is invariant to shifts in this subspace.

To generate $\Theta^\star$, we fix $m, n, r$, let $\Theta_0$ be an $m \times n$ matrix composed of independent draws from a standard normal, take its SVD $\Theta_0 = U \operatorname{diag}(\sigma_1, \sigma_2, \dots) V^T$, truncate it past the top $r$ components $\Theta_1 = U \operatorname{diag}(\sigma_1, \dots, \sigma_r, 0, \dots) V^T$, and renormalize to achieve unit sample standard deviation to get $\Theta^\star$, *i.e.*, $\Theta^\star = \Theta_1/\operatorname{std}(\operatorname{vec}(\Theta_1))$. To generate the choice data, we let $i_t$ be drawn uniformly at random from $\{1, \dots, m\}$, $S_t$ be drawn uniformly at random from all subsets of size 10, and $j_t$ be chosen according to (1) with parameter $\Theta^\star$.

For our estimator we use Algorithm 1 with $\tilde{r} = 2r$, $\lambda = \frac{1}{8}\sqrt{\frac{Kd\log d}{mnT}}$, $\beta_{\text{dec}} = 0.8$, and $\tau = 10^{-10}$. This regularizing coefficient scales with $m, n, d, T$, and $K$ as suggested by Theorem 3.1, but we find the algorithm performs better in practice when we use a smaller constant than that suggested by the theorem.

We plot the results in Figure 1, where error is measured in root mean squared error (RMSE)

$$\operatorname{RMSE}(\Theta) = \sqrt{\operatorname{Avg}\left(\{(\Theta_{ij} - \Theta^\star_{ij})^2\}_{i,j}\right)} = \frac{1}{\sqrt{mn}} \|\Theta - \Theta^\star\|_{\text{F}}.$$

The results show the advantage in efficient use of the data offered by our approach. The results also show that, relative to MLE, the advantage is greatest when the underlying rank $r$ is small and the number of parameters $m \times n$ is large, but that we maintain a significant advantage even for moderate $r$ and $m \times n$. For large numbers of parameters ($m = n \geq 750$), the RMSE of MLE is very large and does not appear in the plots. Only in the case of greatest rank ($r = 25$), smallest number of parameters ($m = n = 100$), and greatest number of observations ($T = 10^6$) does MLE appear to somewhat catch up with our estimator.

In Figure 2, we plot the RMSE of our estimator against the number of observations per type (or item) $T/d$ for a square problem with $d = m = n$. We see nearly the same error curve traced out as we vary the problem size $d$. This scaling shows that our estimator is able to leverage the low-rank assumption and require the same number of choice observations per type to achieve the same RMSE regardless of problem size.

## 7. CONCLUSION

This paper proposes a new model for assortment choices — the low rank MMNL model — in which the preferences of each type follow a parametric multinomial logit distribution, but share a latent linear structure. We show that this preference structure can be efficiently learned both in theory and in practice through a bound on the sample complexity and through numerical simulations. The optimization problem we propose admits a fast algorithm that scales linearly in the number of types and products. The low rank MMNL approach can make learning choice models from choice data significantly more efficient compared with standard methods, and thereby enables a fine-grained understanding of preferences in a diverse population.

## 8. PROOFS OF PRELIMINARY LEMMAS

PROOF (LEMMA 4.1). Let $\Theta e = 0$. Then because $S_t$ is permutation symmetric we get that

$$\mathbb{E}[Y_t(\Theta)] = \frac{1}{m} \left\| \Theta \left( \frac{I_n}{\sqrt{n}} - \frac{\mathbf{1}_{n \times n}}{n\sqrt{n}} \right) \right\|_F^2 = \frac{1}{m}\frac{1}{n} \|\Theta\|_F^2.$$

Therefore, letting $Y'_t(\Theta)$ be an identical and independent replicate of $Y_t(\Theta)$ and letting $\epsilon_t$ be iid Rademacher random variables independent of all else, we have

$$\mathbb{E}\mathcal{M}_{\Gamma,\nu} = \mathbb{E}\left[ \sup_{\Theta \in \mathcal{A}_{\Gamma,\nu}} \frac{1}{T} \sum_{t=1}^T (\mathbb{E}Y'_t(\Theta) - Y_t(\Theta)) \right]$$

$$\leq \mathbb{E}\left[ \sup_{\Theta \in \mathcal{A}_{\Gamma,\nu}} \frac{1}{T} \sum_{t=1}^T (Y'_t(\Theta) - Y_t(\Theta)) \right]$$

$$= \mathbb{E}\left[ \sup_{\Theta \in \mathcal{A}_{\Gamma,\nu}} \frac{1}{T} \sum_{t=1}^T \epsilon_t (Y'_t(\Theta) - Y_t(\Theta)) \right]$$

$$\leq 2\mathbb{E}\left[ \sup_{\Theta \in \mathcal{A}_{\Gamma,\nu}} \frac{1}{T} \sum_{t=1}^T \epsilon_t Y_t(\Theta) \right].$$

Letting $e_S \in \mathbb{R}^n$ be the indicator vector of the set $S$,

$$Y_t(\Theta) = \frac{1}{K_t} \left\| \left( \operatorname{diag}(e_{S_t}) - \frac{1}{K_t} e_{S_t} e_{S_t}^T \right) \Theta^T e_{i_t} \right\|_2^2.$$

Note $\|\cdot\|_2^2$ is $2\gamma k$-Lipschitz with respect to $\infty$-norm on a domain in $[-\gamma, \gamma]^n$ where only $k$ entries are nonzero. Therefore, by Lemma 7 of Bertsimas and Kallus [2014] and by Hölder's inequality, letting $W_t = \sum_{j \in S_t} \epsilon_{tj} \left( e_{i_t} e_j^T - \frac{1}{K_t} \sum_{j' \in S_t} e_{i_t} e_{j'}^T \right)$ where $\epsilon_{tj}$ are iid

Rademacher random variables independent of all else,

$$\mathbb{E}\mathcal{M}_{\Gamma,\nu} \leq 4\gamma \mathbb{E}\left[\sup_{\Theta \in \mathcal{A}_{\Gamma,\nu}} \frac{1}{T}\sum_{t=1}^{T} W_t \cdot \Theta\right]$$

$$\leq 4\gamma \mathbb{E}\left\|\frac{1}{T}\sum_{t=1}^{T} W_t\right\|_2 \sup_{\Theta \in \mathcal{A}_{\Gamma,\nu}} \|\Theta\|_*.$$

Note $\|W_t\|_2 \leq \sqrt{K_t} \leq \sqrt{K}$. Moreover,

$$\mathbb{E}\left[W_t W_t^T | S_t, i_t\right] = (K_t - 1)e_{i_t}e_{i_t}^T,$$

$$\text{and so } \left\|\mathbb{E}\left[W_t W_t^T\right]\right\|_2 = \frac{\mathbb{E}K_t - 1}{m}.$$

Since $S_t|K_t$ is uniform,

$$\mathbb{E}\left[W_t^T W_t | S_t\right] = \text{diag}(e_{S_t}) - \frac{1}{K_t} e_{S_t} e_{S_t}^T,$$

$$\left\|\mathbb{E}\left[W_t^T W_t | K_t\right]\right\|_2 = \frac{K_t - 1}{n - 1},$$

$$\text{and so } \left\|\mathbb{E}\left[W_t^T W_t\right]\right\|_2 \leq \frac{\mathbb{E}K_t - 1}{n - 1},$$

by iterated expectation and Jensen's inequality. The matrix Bernstein inequality [Tropp 2012, Thm. 1.6] gives that $\left\|\frac{1}{T}\sum_{t=1}^T W_t\right\|_2 \geq \delta$ with probability at most

$$(m+n)\max\left\{e^{-\frac{T\delta^2 \min\{m,n-1\}}{4(\mathbb{E}K_t-1)}}, e^{-\frac{\delta}{2\sqrt{K}}}\right\}.$$

Setting the probability to $1/d^{3/2}$ and using $T \leq d^2 \log d$,

$$\mathbb{E}\left[\left\|\frac{1}{T}\sum_{t=1}^T W_t\right\|_2\right] \leq \frac{\sqrt{K}}{d^{3/2}} + 2\sqrt{\frac{(\mathbb{E}K_t - 1)\log(2d^{3/2})}{T\min\{m,n-1\}}}$$

$$\leq \frac{\sqrt{K}}{d^{3/2}} + 2\sqrt{3}\sqrt{\frac{K\log d}{T\min\{m,n-1\}}}$$

$$\leq 5\sqrt{\frac{K\log d}{T\min\{m,n-1\}}} \leq 10\sqrt{\frac{K\log d}{T\min\{m,n\}}}.$$

Putting it all together, we get,

$$\mathbb{E}\mathcal{M}_{\Gamma,\nu} \leq \frac{\nu\Gamma^2}{6\sqrt{mn}}\sqrt{\frac{1}{d\min\{m,n\}}} \leq \frac{\nu}{3}\frac{\Gamma^2}{mn}.$$

Next we use this to prove the concentration of $\mathcal{M}_{\Gamma,\nu}$. Let $\mathcal{M}'_{\Gamma,\nu}$ be a replicate of $\mathcal{M}_{\Gamma,\nu}$ with $i'_t = i_t$, $S'_t = S_t$ for all $t$ except $t'$. The difference $\mathcal{M}_{\Gamma,\nu} - \mathcal{M}'_{\Gamma,\nu}$ is bounded by

$$\frac{1}{T}\sup_{\Theta \in \mathcal{A}_{\Gamma,\nu}}\left(\text{Var}\left(\{\Theta_{i_{t'}j}\}_{j \in S_{t'}}\right) - \text{Var}\left(\{\Theta_{i'_{t'}j}\}_{j \in S'_{t'}}\right)\right) \leq \frac{1}{T}\left(\gamma^2 - 0\right) = \frac{\gamma^2}{T}.$$

Hence, by McDiarmid's inequality, we have

$$\mathbb{P}\left(\mathcal{M}_{\Gamma,\nu} - \mathbb{E}\mathcal{M}_{\Gamma,\nu} \geq \delta\right) \leq e^{-\frac{2T\delta^2}{\gamma^4}}.$$

Using $\delta = \frac{2\nu}{3}\frac{\Gamma^2}{mn}$ and $\mathbb{E}\mathcal{M}_{\Gamma,\nu} \leq \delta/2$ we get the result. $\square$

PROOF (LEMMA 4.2). Since $\|\cdot\|_* \geq \|\cdot\|_F$, we have $\inf_{\Theta \in \mathcal{A}^*} \|\Theta\|_F \geq \tau := 128\sqrt{Kmn}\gamma\sqrt{d\log d/T}$. Let $\mathcal{A}_l = \mathcal{A}^* \cap \{\sqrt{2}^{l-1} \leq \|\Theta\|_F \leq \sqrt{2}^l\}$ and note that $\mathcal{A}^* = \bigcup_{l=1}^\infty \mathcal{A}_l$ and $\mathcal{A}_l \subset \mathcal{A}_{\sqrt{2}^l\tau, 1/4}$. Moreover, if $\Theta \in \mathcal{A}_l$ has $\frac{1}{T}\sum_{t=1}^T Y_t(\Theta) < \frac{1}{2mn}\|\Theta\|_F^2$ then

$$\frac{\|\Theta\|_F^2}{mn} - \frac{1}{T}\sum_{t=1}^T Y_t(\Theta) > \frac{\|\Theta\|_F^2}{2mn} \geq \frac{1}{mn}\frac{1}{4}(\sqrt{2}^l\tau)^2.$$

Therefore, with $p$ denoting the probability in the statement of the theorem to be bounded,

$$
\begin{aligned}
1 - p &\leq \min\left\{1, \sum_{l=1}^\infty \mathbb{P}\left(\mathcal{M}_{\sqrt{2}^l\tau, 1/4} > \frac{1}{4}\frac{1}{mn}\left(\sqrt{2}^l\tau\right)^2\right)\right\} \\
&\leq \min\left\{1, \sum_{l=1}^\infty \exp\left(-\frac{1}{18}\frac{4^l\tau^4 T}{m^2n^2\gamma^4}\right)\right\} \\
&\leq \min\left\{1, \sum_{l=1}^\infty \exp\left(-\frac{2}{9}\frac{\tau^4 T}{m^2n^2\gamma^4}l\right)\right\} \\
&= \min\left\{1, \left(\exp\left(\frac{1}{72}\frac{\tau^4 T}{m^2n^2\gamma^4}\right) - 1\right)^{-1}\right\} \\
&\leq 2\exp\left(-\frac{2}{9}\frac{\tau^4 T}{m^2n^2\gamma^4}\right) \\
&= 2\exp\left(-59652323.6 d^2 K^2 (\log d)^2/T\right) \\
&\leq 2d^{-59652323.6\cdot 2^2} \\
&\leq 2d^{-238609294} \\
&\leq 2d^{-2^{27}},
\end{aligned}
$$

using Lemma 4.1, $T \leq d^2 \log d$, and $K \geq 2$. $\square$

PROOF (LEMMA 4.3). Let $R_t(\Theta) = e_{i_t}e_{j_t} - \frac{\sum_{j\in S_t}\eta_{tj}(\Theta)e_{i_t}e_j^T}{\sum_{j\in S_t}\eta_{tj}(\Theta)}$ where $\eta_{tj}(\Theta) = e^{-\Theta_{i_tj}}$. Then $\nabla L_T(\Theta) = \frac{1}{T}\sum_{t=1}^T R_t(\Theta)$. Note that because $j_t$ is drawn according to $\Theta^*$, we have that $\mathbb{E}\left[R_t(\Theta^*)|i_t, S_t\right] = 0$ and hence $\mathbb{E}R_t(\Theta^*) = 0$. Let $R_t = R_t(\Theta^*)$, $\eta_{tj} = \eta_{tj}(\Theta^*)$. Note that $\|R_t\|_2 \leq \sqrt{2}$. Moreover, letting $\kappa_t = \sum_{l\in S_t}\eta_{tl}$,

$$R_t R_t^T = e_{i_t}e_{i_t}\left(1 - \frac{2\eta_{tj_t}}{\kappa_t} + \frac{\sum_{j\in S_t}\eta_{tj}^2}{\kappa_t^2}\right).$$

Since by Jensen's inequality the multiplier in the parentheses is no greater than 2, we get $\left|\left|\mathbb{E}\left[R_t R_t^T\right]\right|\right|_2 \leq \frac{2}{m} \leq \frac{2K}{m}$. Letting $y_{tj} = \mathbb{I}[j = j_t]$, we have

$$R_t^T R_t = \sum_{j,k \in S_t} e_j e_k^T \left( y_{tj} y_{tk} - \frac{2 y_{tj} \eta_{tk}}{\kappa_t} + \frac{\eta_{tj} \eta_{tk}}{\kappa_t^2} \right).$$

Since $y_{tj} \geq 0$, $\eta_{tj} \geq 0$, and $y_{tj} y_{tk} \leq \mathbb{I}[j = k]$,

$$\left|\left|\mathbb{E}\left[R_t^T R_t\right]\right|\right|_2 \leq \left|\left|\mathbb{E}\left[\mathrm{diag}(e_{S_t})\right]\right|\right|_2 + \left|\left|\frac{\mathbb{E}\left[e_{S_t} e_{S_t}^T\right]}{K_t^2}\right|\right|_2$$
$$\leq \frac{K}{n} + \frac{1}{n} \leq \frac{2K}{n}.$$

By matrix Bernstein inequality [Tropp 2012, Thm. 1.6], $\left|\left| \frac{1}{T} \sum_{t=1}^T R_t \right|\right|_2 \geq \delta$ with probability at most

$$2d \max \left\{ e^{-\frac{T \delta^2 \min\{m,n\}}{8K}}, e^{-\frac{T\delta}{2\sqrt{2}}} \right\}.$$

Hence, with probability at least $1 - 2d^{-3}$,

$$\left|\left| \frac{1}{T} \sum_{t=1}^T R_t \right|\right|_2 \leq \sqrt{\frac{32 K \log d}{\min\{m,n\} T}} \leq \sqrt{\frac{128 K d \log d}{mnT}}.$$

□